\documentclass[%
 reprint,
 amsmath,amssymb,
 aps,
 linenumbers,
]{revtex4-2}

\usepackage{graphicx}
\usepackage{dcolumn}
\usepackage{bm}
\usepackage{hyperref}

\usepackage{float}
\usepackage{makecell}
\usepackage{enumitem}
\usepackage{comment}
\usepackage{color}
\usepackage{lineno}

\newcommand{\pasteuraffil}{Institut Pasteur, Universit\'e Paris Cit\'e, CNRS UMR 3571, Decision and Bayesian Computation, 75015 Paris, France.}

\newcommand{\Prob}{\mathbb{P}}

\newcommand{\arm}{i}
\newcommand{\arma}[1]{\arm_{#1}}
\newcommand{\armt}{\arm_{t}}
\newcommand{\Arm}{A}
\newcommand{\Reward}{X}
\newcommand{\Armt}{\Arm_{t}}
\newcommand{\Rewardt}{\Reward_t}
\newcommand{\Rewardtau}{\Reward_\tau}

\newcommand{\arms}{\mathbb{A}}

\newcommand{\Regret}{{R}(t)} 
\newcommand{\Stail}{\tilde{S}_{\textrm{tail}}}
\newcommand{\Stailtheory}{S_{\textrm{tail}}}

\newcommand{\Sc}{\tilde{S}_{\mathrm{body}}}
\newcommand{\Sctheory}{S_{\mathrm{body}}}

\newcommand{\Smax}{S}
\newcommand{\Sapp}{\tilde{\Smax}}
\newcommand{\pmax}{p_{\max}}
\newcommand{\ptail}{c_{\mathrm{tail}}}

\newcommand{\Sp}{S_{\mathrm{p}}}

\newcommand{\barm}{\mu^*}
\newcommand{\marm}{\mu}
\newcommand{\marmi}[1]{\mu_{#1}}

\newcommand{\psup}[2]{p_{#1 >  #2}}
\newcommand{\psupmaxmin}{p_{\meana{\maxa} > \meana{\mina}}}
\newcommand{\psupminmax}{p_{\meana{\mina} > \meana{\maxa}}}

\newcommand{\BetaIncS}[1]{I_{#1}}

\newcommand{\BetaDisLaw}[2]{X_{\mathcal{B}(#1+1,#2+1)}}

\newcommand{\erf}{\mathrm{erf}}
\newcommand{\erfc}{\mathrm{erfc}}
\newcommand{\kull}{K_{\mathbb{B}}}

\newcommand{\ndraw}[1]{n_{#1}}
\newcommand{\ndrawt}[1]{n_{#1}(t)}
\newcommand{\rew}[1]{r_{#1}}
\newcommand{\rewt}[1]{r_{#1}(t)}

\newcommand{\teq}{\tilde{\theta}_{\mathrm{eq}}}
\newcommand{\tp}{\theta_{\mathrm{eq}}}

\newcommand{\mean}{\theta}
\newcommand{\meana}[1]{\theta_{#1}}

\newcommand{\armmin}{\arm_{\mina}}
\newcommand{\meanmax}{\mean_{\maxa}}
\newcommand{\meanmin}{\mean_{\mina}}
\newcommand{\renormmax}{\Theta_{\maxa}}
\newcommand{\renormmin}{\Theta_{\mina}}

\newcommand{\N}[1]{N_{#1}}
\newcommand{\V}[1]{V_{#1}}

\newcommand{\Deltam}{\Delta}
\newcommand{\Vtot}{V_{\mathrm{t}}}

\newcommand{\Vmax}{V_{\mathrm{max}}}
\newcommand{\Vmin}{V_{\mathrm{min}}}
\newcommand{\maxa}{\mathrm{max}}
\newcommand{\mina}{\mathrm{min}}

\newcommand{\nmin}{\hat{N}_{\mathrm{min}}}
\newcommand{\mmin}{\hat{\theta}_{\mathrm{min}}}

\newcommand{\parmax}{p_{\mathrm{\theta}_{\maxa}}}
\newcommand{\parmin}{p_{\mathrm{\theta}_{\mina}}}

\newcommand{\binf}{\mean_{\mathrm{inf}}}
\newcommand{\bsup}{\mean_{\mathrm{sup}}}
\newcommand{\thetaDomain}{\Theta}

\newcommand{\kulleib}{Kullback-Leibler }

\newcommand{\gradsp}{\Delta_{\mathrm{max}, \mathrm{min}} S_{\mathrm{p}} }

\newcommand{\Ac}{A_{\mathrm{c}}}
\newcommand{\Acval}{1.25889}

\newcommand{\egreedy}{$\epsilon$-greedy }
\newcommand{\engreedy}{$\epsilon_{\mathrm{n}}$-greedy }

\newcommand{\algoname}{AIM }
\newcommand{\algonamese}{AIM}

\begin{document}

\preprint{APS/123-QED}
\nolinenumbers
\title{Approximate information for efficient exploration-exploitation strategies}

\author{Alex Barbier--Chebbah }
\email{alex.barbier-chebbah@pasteur.fr}
\author{Christian L.\ Vestergaard}
\affiliation{\pasteuraffil}
\author{Jean-Baptiste Masson}
\affiliation{\pasteuraffil}

\date{\today}

\begin{abstract}
This paper addresses the exploration-exploitation dilemma inherent in decision-making, focusing on multi-armed bandit problems. The problems involve an agent deciding whether to exploit current knowledge for immediate gains or explore new avenues for potential long-term rewards. 
We here introduce a novel algorithm, approximate information maximization (\algonamese), which employs an analytical approximation of the entropy gradient to choose which arm to pull at each point in time.
\algoname matches the performance of Infomax and Thompson sampling while also offering enhanced computational speed, determinism, and tractability. 
Empirical evaluation of \algoname indicates its compliance with the Lai \& Robbins asymptotic bound and demonstrates its robustness for a range of priors. 
Its expression is tunable, which allows for specific optimization in various settings.
\end{abstract}

\maketitle

\paragraph*{\label{sec:intro}Introduction.}
The exploration-exploitation dilemma is a fundamental challenge in decision-making. It arises when an agent must choose between exploiting its current knowledge to maximize immediate rewards or acquiring new information that may lead to greater long-term gains. This dilemma is ubiquitous in various fields, from anomaly detection~\cite{ding_interactive_2019} to the modelling of biological search strategies~\cite{vergassola_infotaxis_2007, martinez_using_2014, carde_navigation_2021} and human decision-making~\cite{cohen_should_2007,hills_exploration_2015,mehlhorn_unpacking_2015,doya_bayesian_2007, jepma_pupil_2011}.

The multi-armed bandit problem is a paradigmatic example of an explore-exploit problem and has been extensively studied and applied in a range of fields, including applied mathematics~\cite{slivkins_introduction_2019,gittins_bandit_1979,zhou_survey_2016,bubeck_pure_2011,bayati_unreasonable_2020,bouneffouf_survey_2020,auer_finite-time_2002} to animal behavior~\cite{morimoto_foraging_2019}, neuroscience~\cite{wilson_balancing_2021,tervo_behavioral_2014,bouneffouf_bandit_2017,markovic_empirical_2021}, clinical trials~\cite{durand_contextual_2018,villar_bandits_2018,villar_multi-armed_2015}, finance~\cite{shen_portfolio_2015}, epidemic control~\cite{lin_optimal_2022}, and reinforcement-learning~\cite{silver_mastering_2016,ryzhov_knowledge_2012}, among others.
In the multi-armed bandit problem, an agent is presented with a set of possible actions, or "arms", each associated with a probabilistic reward (akin to a multi-armed slot machines game).  
The agent must choose which arm to pull at each time step to maximize its cumulative reward over a fixed or infinite time horizon. 
Hence, at each time step, the agent can either play the arm with the better rewards to improve the knowledge on that arm or explore new arms to test if they would not lead to increased rewards.

\begin{figure}
\centering
\includegraphics[scale=0.4]{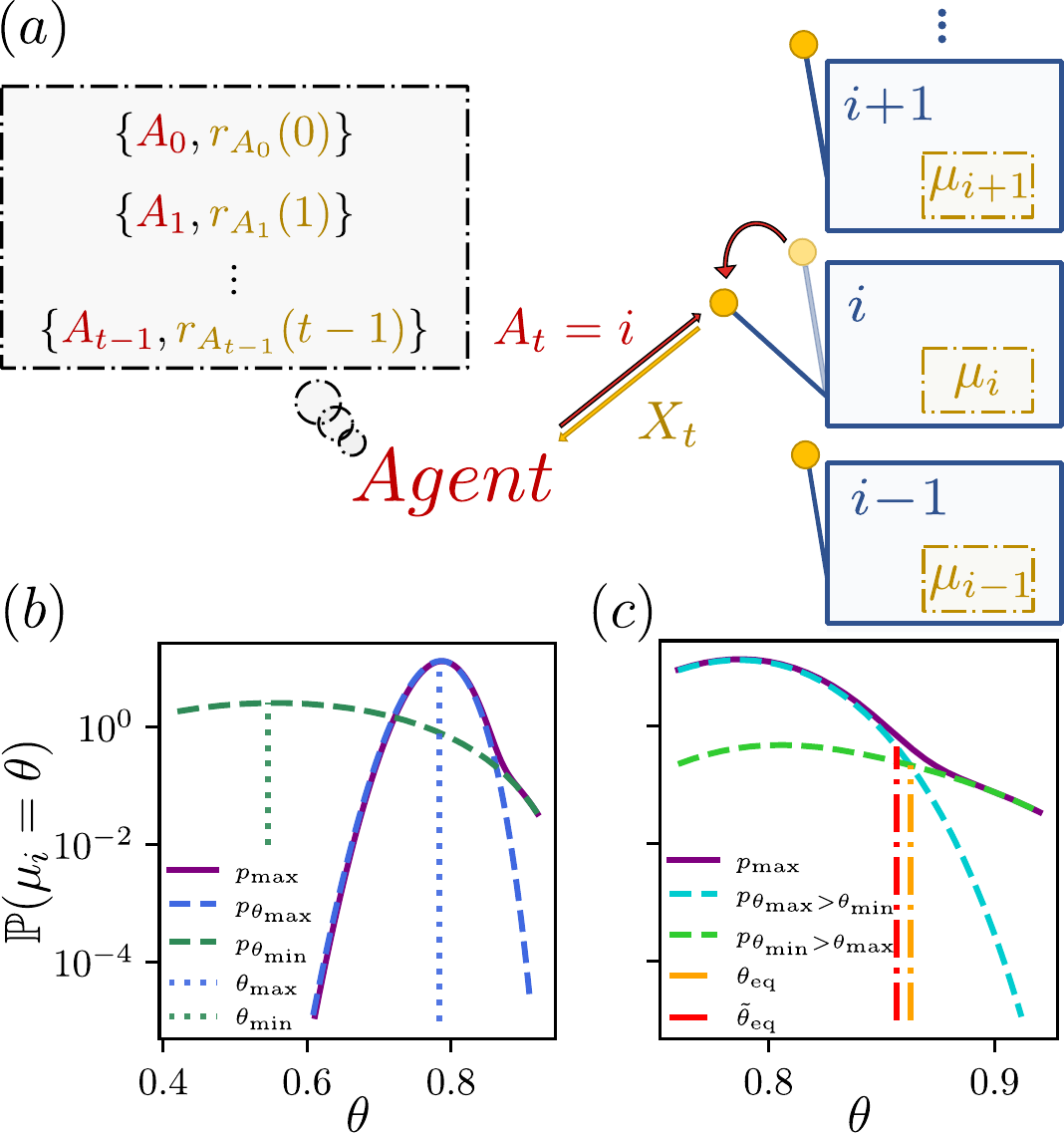}
\caption{\textbf{(a)} Illustration of the multi-armed bandit problem. At each time step $t$ the agent chooses an action $i=A_t$ that returns a reward $\rewt{i}$ drawn from a distribution of unknown mean $\marmi{i}$. The agent's goal is to minimize the cumulative regret $R(t)$ [see Eq.~\eqref{regretdef}]. \textbf{(b)} Posterior distributions of bandit values after playing the 2-armed Bernoulli game with $\rewt{1}=5$,  $\ndrawt{1}=9$, $\rewt{2}=41$, $\ndrawt{2}=192$,  where $\rewt{i}$, $\ndrawt{i}$ are respectively the cumulative reward and number of draws of arm $i$. In blue, the posterior distribution, $\parmax$,  of the reward of the current best arm. Vertical green and blue lines are the current average rewards of the suboptimal (denoted $\meanmin$) and optimal arm ($\meanmax$). In green, the posterior distribution, $\parmin$, of the current sub-optimal arm. In purple, the posterior distribution, $\pmax$, of the maximum reward of all arms. \textbf{(c)} Zoomed plot of \textbf{(b)} in the region where the posterior distribution of the maximal reward value transition from being dominated by $\psupmaxmin$ to being dominated by $\psupminmax$. In purple  $\pmax$. In light blue, the probability, $\psupmaxmin$, is that the optimal arm's gain is superior to the suboptimal arm. In light green, the probability, $\psupminmax$, is that the gain of the suboptimal arm is superior to that of the optimal arm. The orange vertical line is the transition value $\tp$ and the red vertical line its approximation $\teq$ (see Supplemental Material \ref{SiStail1} \& \ref{refteqapprox} for its derivation).} 
\label{fig:overview}
\end{figure}

In the following we begin with a brief introduction to the bandit problem, followed by a presentation of our novel approximate information procedure, completed with its corresponding analytical expression. We then provide empirical evidence of the procedure's efficacy before delving into a discussion of its various properties and implications.

We consider the classic multi-armed bandit setting~\cite{sutton_reinforcement_1998}. 
At each point in time, $t$, an agent chooses an arm, $\Arm_t$, between $K$ different arms, $\arms =  \{1, 2, \dots, K\}$. The chosen arm, $\armt$ returns a stochastic reward, $\Rewardt$, drawn from a distribution whose mean, $\marmi{\armt}$, is unknown to the agent [Fig.~\ref{fig:overview}(a)]. 
The agent's goal is to maximize the cumulative reward (equivalently, minimize the cumulative regret) with no time horizon. Formally, we aim to minimize the expected regret \cite{sutton_reinforcement_1998}, $\mathbb{E}[\Regret]$, with
\begin{equation}\label{regretdef}
    \Regret = \barm t - \sum_{\tau=1}^t \Rewardtau .
\end{equation}
The regret, $\Regret$, measures the cumulative difference between the rewards obtained by the algorithm and the expected reward that it would have obtained by choosing the best action. 
Optimal strategies, regardless of their details, are characterized by the following asymptotic bound (the Lai and Robbins bound)~\cite{lai_asymptotically_1985}: 
\begin{equation}\label{eq:LR_bound}
	\langle R(t) \rangle_{t\rightarrow\infty}\geq \beta\, \log(t),
\end{equation}
where $\beta$ is a constant factor that depends on the reward distributions.

\begin{figure*}[htbp!]
\includegraphics[scale=0.5]{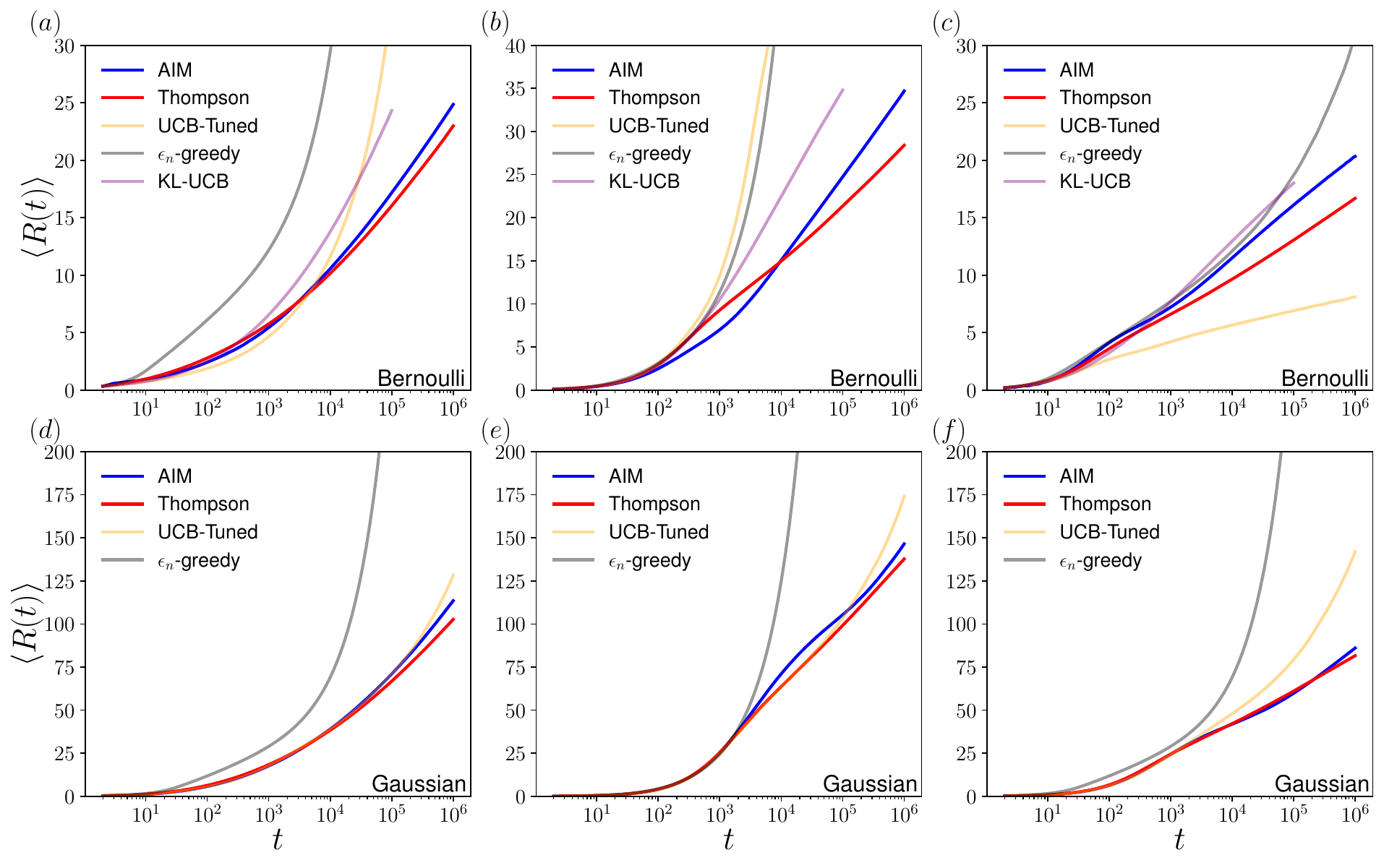}
\caption{Temporal evolution of the regret for Bernoulli \textbf{(a-c)} and Gaussian \textbf{(d-f)} 2-armed bandits. 
In blue \algonamese, in red Thompson sampling, in yellow UCB-tuned, in grey $\epsilon_{n}$-greedy, and in purple KL-UCB. Details of simulations and the tuning required for some algorithms are provided in Supplemental Material \ref{SiAIMimp} and \ref{Simethodsalgo}. 
True parameters were drawn uniformly in $]0,1[$ for both bandits in \textbf{(a,d)}, parameters were set to $\marmi{1}=0.7$, $\marmi{1}=0.8$ for \textbf{(b,e)} and to $\marmi{1}=0.1$, $\marmi{1}=0.3$ for \textbf{(c,f)}.}
\label{fig:2}
\end{figure*}

Multiple strategies attain the Lai and Robbins bound [Eq.~\eqref{eq:LR_bound}]. 
Notably, the $\epsilon_{n}$-greedy strategy~\cite{slivkins_introduction_2019}, which plays the best current arm with probability $1-\epsilon_n$ and randomly samples other arms with probability $\epsilon_n$, with a time-varying $\epsilon_{n}$; 
the Upper Confidence Bound-2 (UCB-2) algorithm~\cite{auer_finite-time_2002}, which relies on a tuned confidence index associated to each arm to decide which arm to play; 
Thompson sampling (proportional betting), which relies on sampling the action from the posterior distribution that it maximizes the expected reward. 
Importantly, methods such as the $\epsilon_{n}$-greedy and UCB-based algorithms require parameter tuning to reach the Lai and Robbins bound, making them sensitive to uncertainties and variations of the prior information used for tuning.

\paragraph*{Approximate information maximization for bandit problems.}
We aim here to develop a tractable, functional-based algorithm for the multi-armed bandit problem. 
Inspired by the Infomax principle~\cite{reddy_infomax_2016, vergassola_infotaxis_2007}, we rely on the entropy as a functional to optimise to decide which arm to play. 
Contrary to classical bandit algorithms, the entropy encompasses the information carried by all arms in a single functional, thus characterising the global state of the game. More precisely, we aim to optimise $\Smax$, the entropy of the posterior distribution of the value of the maximal reward, $\pmax$,
\begin{equation}\label{Smax}
 \Smax = -\int_{\thetaDomain} \: \pmax(\mean) \ln \pmax(\mean) d\mean,
\end{equation}
where $\thetaDomain = [\binf, \bsup]$ is the support of $\pmax$ (which depends on the nature of the game), and
\begin{equation}\label{pmaxgeneralexpression}
\begin{split}
\pmax(\mean) &= \sum_{\arm=0}^K \Prob(\marmi{i} = \mean) \prod_{j\neq i} \Prob(\marmi{j} \leq \mean).
\end{split}
\end{equation}

The entropy $\Smax$ summarizes the information about the state of the game and we require our algorithm to greedily optimise its gradient, i.e., to select the next arm according to:
\begin{equation}\label{greedygradient}
  \underset{i=1..K}{\mathrm{argmin}} \langle {\Smax(t+1)} - \Smax(t) | \Arm_{t+1} = \arm \rangle .
\end{equation}
By doing so, the algorithm seeks to maximize the expected decrease in entropy, conditioned on the current knowledge of the game. This strategy has shown to be competitive with state-of-the-art algorithms and attain the Lai and Robbins bound~\cite{reddy_infomax_2016}.

However, while Eq.~\eqref{greedygradient} can be numerically evaluated, it cannot be computed in closed form for most bandit problems. To obtain an algorithm that is both tractable and computationally efficient, a second functional approximating the entropy has to be derived. 

Hence, we devise a set of approximations of both $\pmax$ and $\Smax$ to get a tractable algorithm. We develop our approach on the 2-armed bandit. We denote the arms according to their current mean rewards, respectively the maximum one by $\arm_{\maxa}$ (with expected reward $\mean_{\maxa}$) and the minimum by $\arm_{\mina}$ (with $\mean_{\mina}$). 
Note that the true expected reward of $\arm_{\maxa}$ may be smaller than that of $\arm_{\mina}$ due to the stochasticity of the game. 

Our approximate form of the entropy reads: 
\begin{equation}\label{theoryS}
\begin{split}
\Sapp &= (1 - \ptail) \Sc + \Stail - (1-\ptail) \ln(1- \ptail).
\end{split}
\end{equation}
It decomposes the entropy into three tractable terms corresponding to approximations made on $\pmax$. 
The first term, $\Sc$, approximates the entropy of the mode of $\pmax$. The second, $\Stail$, captures the entropy of the tail (on the high reward side, see Fig.~\ref{fig:overview}(b,c)) of $\pmax$. These approximate entropies are weighted by factors depending on $\ptail$, a corrective term that compensates for an extension of the integral boundaries in order to make the entropy evaluations analytically tractable (see Supplemental Material \ref{SiStail1}  for details). 

More precisely, the tail term reads: 
\begin{equation}\label{Stailexp}
\Stail = - \int_{\teq}^{\bsup} \parmin(\mean) \ln \parmin(\mean) d\mean,
\end{equation}
where $\teq$ is the approximation of $\tp$, the value of  $\theta$ where the probability of being the maximum is identical for both arms  (see red and orange curves on Fig.~\ref{fig:overview}(c)), and $\parmin(\mean) = \Prob(\marmi{\armmin} = \mean)$ is the posterior probability of the current suboptimal arm having expected reward $\theta$. 

The approximate entropy of the main mode is split into two terms: 
\begin{equation}\label{Sbodyexp}
\begin{split}
\Sc &= - \int_{\thetaDomain} \psupmaxmin(\mean)  \ln\parmax(\mean) d\theta \\
&\quad - \Ac \int_{\thetaDomain} \psupminmax(\mean) d\theta , 
\end{split}
\end{equation}
where $\parmax(\mean)$ is the posterior probability at $\mean$ of the current optimal arm, 
$\psup{\meana{i}}{\meana{j}}(\mean) = \Prob(\marmi{i} = \mean, \marmi{i} \geq \marmi{j} )$ is the  posterior probability for the expected reward $\mean$ of arm $i$ to be larger than $\meana{j}$, 
and $\Ac=\Acval$ is a predetermined constant [see Eq.~\eqref{SiSb8} in Supplemental Material~\ref{SiAc}].
The first term in Eq.~\eqref{Sbodyexp} is the leading-order term  of the mode of $\pmax$, dominated by the current optimal arm,  whereas the second term handles the corrections induced by the suboptimal arm in the vicinity of $\meanmax$ (see Supplemental Material \ref{SiAc} for details).

Finally, the third, corrective term  in Eq.~\eqref{Stailexp} is $\ptail = \int_{\teq}^{\bsup} \parmin(\mean) d\mean$. 

We propose {approximate information maximization} (\algonamese), an algorithm that consists in evaluating Eq.~\eqref{theoryS} for each arm in each time step $t$ and choosing the one that minimizes the expected value of $\Sapp(t+1)$ according to Eq.~\eqref{greedygradient}. Depending on the reward distributions, and their associated $\thetaDomain$, the log dependencies inside $\Sc$ and $\Stail$ can be integrated analytically or approximated by its long-time asymptote (see Supplemental Material \ref{analytic_derivation} for a detailed deviation of all terms). 
Then, \algoname provides a direct implementation following an analytically tractable  expression.

\begin{figure}
\centering
\includegraphics[scale=0.37]{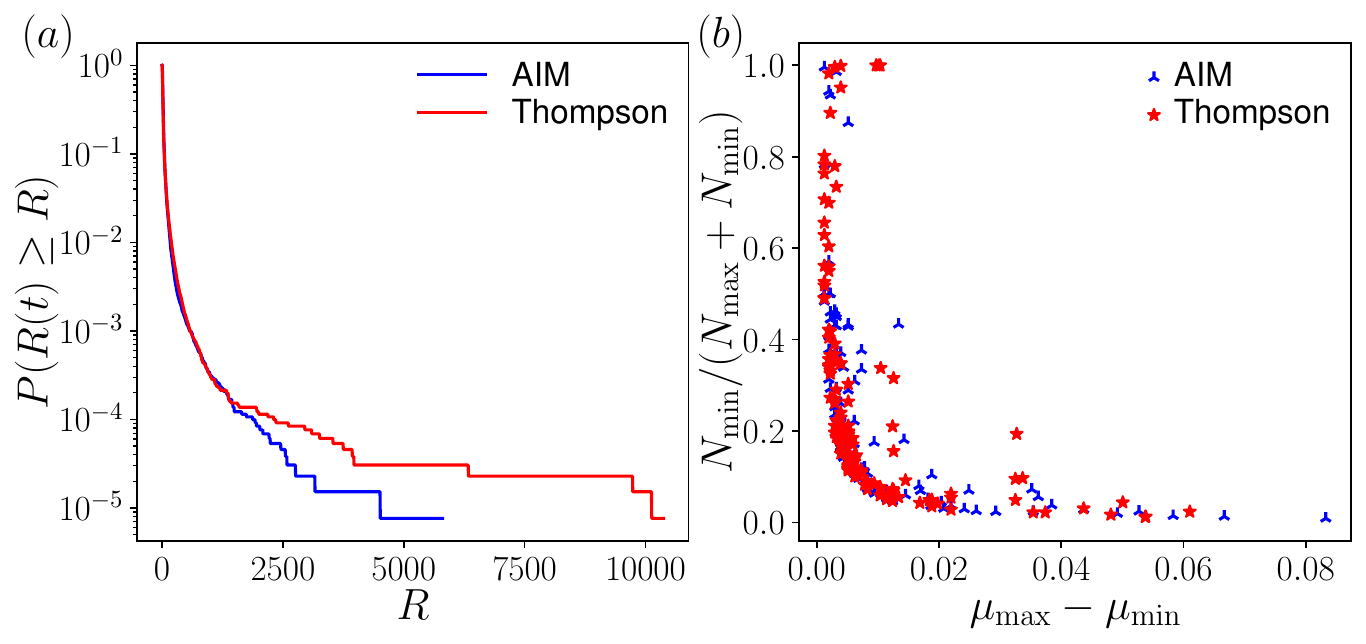}
\caption{Regret distribution and rare events. \textbf{(a)} The probability of obtaining a cumulative regret superior to $R$ for both Thompson and \algoname playing Bernoulli games with uniform priors and $N=2^{17}$ realizations). \algoname has an exponential decay similar to Thompson algorithm. \textbf{(b)} Fraction of the sub-dominant arm, drawn for high regret events ($0.1\%$) and plotted  along the mean difference $\mu_{\maxa}-\mu_{\mina}$. In both \textbf{(a-b)} Thompson and \algoname exhibit the same behaviour.}
\label{fig:3}
\end{figure}


\paragraph*{Results.}
We demonstrate the performance of \algoname on the paradigmatic Bernoulli bandits~\cite{slivkins_introduction_2019,pilarski_optimal_2021,thompson_likelihood_1933} and on Gaussian bandits~\cite{honda_asymptotically_2010} with unknown mean $\marmi{\arm}\in[0,1]$ and unit variance. Supplementary Table~\ref{table:maintable} lists analytic expressions for the  terms of $\Sapp$ [Eq.~\eqref{theoryS}] for each problem. 


Figure~\ref{fig:2} compares the performance of the AIM algorithm with other state-of-the-art algorithms on numerically generated data (see Supplemental Material \ref{SiAIMimp} \& \ref{Simethodsalgo} for implementation of \algoname and other classic bandit strategies).
For both Bernoulli and Gaussian bandits, \algoname empirically follows the Lai \& Robins bound, with a regret scaling as $\log\left(t\right)$. Its long time performance matches that of Thompson sampling while relying on a simple analytical formula. Additionally, \algoname outperforms Thompson methods at intermediate times for challenging parameter configurations [Fig.~\ref{fig:2}(b)].

The following heuristic argument qualifies the optimal asymptotic scaling of \algonamese.
Assuming $t \gg 1$ and $\N{\maxa} \gg \N{\mina} \gg 1$, i.e., the best arm has been predominantly pulled. 
Then, the variation along $\N{\mina}$ and $\N{\maxa} = t-\N{\mina}$ of the approximate entropy reads:
%
%
%
%
%
\begin{equation}\label{gradientSNmin}
\begin{split}
\frac{\partial \Sapp}{\partial \N{\mina} } &=   (1-\ptail) \frac{\partial \Sc}{\partial \N{\mina} } + \frac{\partial \Stail}{\partial \N{\mina} }  +.. \\&   (-\Sc  + \ln(1-\ptail) + 1) \frac{\partial \ptail}{\partial \N{\mina} },
\end{split}
\end{equation}

To leading order, the minimum of Eq.~\eqref{gradientSNmin} is found at $\N{\mina} \sim \ln(t)/\kull(\marmi{\mina}, \marmi{\maxa})$. For Bernoulli bandits, where $\kull(\marmi{\mina}, \marmi{\maxa})$ is the Kullback-Leibler divergence between the reward distributions, thus recovering the Lai and Robbins bound (see derivation in Supplemental Material \ref{refasymptotepart}). 
Note that this derivation is not entirely rigorous as it assumes that, after a certain time, we can be sure that the optimal arm has been predominantly pulled. We checked this assumption by investigating the asymptotic behaviour of high cumulative regret events (Fig.~\ref{fig:3}), for which the sub-dominant arm has been drawn a non-negligible fraction of time. These events are exponentially rare and happen only for small $\marmi{\maxa} - \marmi{\mina}$, which require exponentially long times to be distinguished (a behaviour that is shared by Thompson sampling).

\paragraph*{Conclusion.}
In this study, we present a new approach, \algonamese, designed to effectively balance exploration and exploitation in multi-armed bandit problems. 
\algoname employs an analytic approximation of the entropy gradient to select the optimal arm. This novel approach mirrors the performance of Infomax (see Supplemental Material \ref{Simethodsalgo} and Fig.~\ref{fig:2sup}), from which it is derived, while offering improved computational speed. It also parallels Thompson sampling in functionality, yet outperforms it in terms of being deterministic and more easily managed.

Empirical testing demonstrated that \algoname complies with the Lai and Robbins bound and exhibits robustness to a broad spectrum of priors. Furthermore, since it relies on an analytic expression, \algoname can easily be fine-tuned to optimise performance in various scenarios, while still satisfying the Lai and Robbins bounds. 
Specifically, tuned \algoname is highly efficient for K-armed bandits with $K>2$ (see Supplemental Material \ref{SItuned} and Fig.~\ref{fig:tuned} for derivation and examples).

Due to its reliance on a single, analytically tractable functional expression, \algoname proves adaptable for different bandit problems, particularly where other approaches may face efficiency constraints. 
Interesting future research directions include devising a rigorous proof of optimality, applying and optimising \algoname to multi-armed problems with finite horizons, with insufficient time to sample all bandits, and its extension to Monte-Carlo path-planning schemes.



\paragraph*{Acknowledgments.}
We thank Etienne Boursier for helpful discussions for optimality of \algonamese.


\begin{titlepage}
  \vspace{1cm}\.\\
  \centering{\Large{{\bf Supplemental material}}}
  \clearpage
\end{titlepage}

\setcounter{figure}{0}
\setcounter{equation}{0}
\renewcommand{\figurename}{{SUPPLEMENTARY FIG.}}
\renewcommand{\thefigure}{{S\arabic{figure}}}
\renewcommand{\thesection}{{S\arabic{section}}}
\renewcommand{\theequation}{{S\arabic{equation}}}
\renewcommand{\tablename}{{SUPPLEMENTARY TABLE}}
\renewcommand{\thetable}{{S\arabic{table}}}

\begin{widetext}

\section{Entropy approximation}\label{Siapproximation}

Here, we derive the approximations $\Stail$ and $\Sc$  constituting Eq.~\eqref{theoryS} in the main text.
Equation~\eqref{theoryS} relies on the observation that that functional form of the posterior $\pmax$ can naturally be split in two distinct parts above and below the point $\teq\approx\tp$,
\begin{equation}\label{SiSsplit}
    \Smax = \Sctheory + \Stailtheory ,
\end{equation}
with
\begin{equation}\label{SiScStail}
    \Sctheory = -\int_{\binf}^{\teq} \pmax(\mean) \ln \pmax(\mean) d\mean\,,\quad 
    \Stailtheory = -\int_{\teq}^{\bsup} \pmax(\mean) \ln \pmax(\mean) d\mean .
\end{equation}
The individual contribution, $\Sctheory$ and $\Stailtheory$ are easier to approximate using standard techniques than the full expression, $\Smax$. 
We detail these approximations below.

\subsection{Approximation of the main mode's contribution }\label{SiAc}


The approximations leading to $\Sc$ derives from decomposing $\Sctheory$ as:
\begin{equation}\label{SiSapp1}
\begin{split}
\Sctheory &=   -\int_{\binf}^{\teq} \pmax \ln\left(\psupmaxmin \right)  d\theta -\int_{\binf}^{\teq}  \pmax \ln\left( 1 +\frac{ \psupminmax}{\psupmaxmin}\right) d\theta.\\
\end{split}
\end{equation}

To be able to perform the integration analytically, we extend the upper bound of the integrals from $\teq$ to $\bsup$. 
This requires neglecting the contribution from $\psup{\meana{\mina}}{\meana{\maxa}}$ in the first term, resulting in the weight normalisation factor $\ptail$ appearing in Eq.~\eqref{theoryS}. 
We furthermore approximate $\ln(\psupmaxmin )$ by $\ln(\parmax)$ in the first term. 
Next, we approximate the second term of Eq.~\eqref{SiSapp1} by
\begin{equation}\label{SiSb7}
\ln\left( 1 +\frac{ \psupminmax}{\psupmaxmin}\right) \approx   \Ac \frac{\psupminmax}{\psupmaxmin + \psupminmax} ,
\end{equation}
which is a variation of an approximation deduced from the Taylor series of the inverse hyperbolic tangent for  $0 <  \frac{ \psupminmax}{\psupmaxmin} < 1$. 
First, note that this term should contribute significantly only when  $\frac{ \psupminmax}{\psupmaxmin} < 1$ since the entropy has already been partitioned. Thus, Eq.~\ref{SiSb7} choice is justified because it stays bounded even for $\psupminmax \gg \psupmaxmin $ which occurs since the integral bounds have been pushed above $\teq$. Finally, $\Ac$ is obtained as the solution to:
\begin{equation}\label{SiSb8}
 \int_0^1 \ln(1 +x ) =  \Ac  \int_0^1 \frac{x}{x+1} ,
\end{equation}
leading to $\Ac=\Acval$.
Taken altogether, this leads to Eq.~\eqref{theoryS}. 

\subsection{Approximation of the tail contribution }\label{SiStail1}


The approximation expression for the tail contribution to the entropy, $\Stail$ [Eq.~\eqref{Stailexp}] is obtained from $\Stail$ [Eq.~\eqref{SiScStail}] by neglecting the contribution from $\arm_{\maxa}$, i.e., approximating $\pmax$ by $\parmin$.  
This approximation requires our body-tail separator to precisely determine when the best arm contribution becomes sub-dominant. 
Rephrased differently, $\teq$ approximates the transition value $\tp$ where the current less expected arm will become more likely to be the maximum than the best expected arm. At long times, since $\arm_{\maxa}$ must be much more selected than the suboptimal one, we should observe a distribution $\parmax$ that is highly contracted compared to $\parmin$. This effect will result in a tail that is mostly dominated by $\parmin$ justifying our previous assumption.

\section{Analytical derivation of \algoname } \label{analytic_derivation}

Here, we summarize all the steps leading to the analytic expressions used in \algoname for 2-arms study case and exhibited in Supplementary Table~\ref{table:maintable}.  


\begin{table*}[htbp]
\centering
\begin{tabular}{ |c |c |c|}
\hline 
 & Bernoulli reward &  Gaussian reward \rule[0pt]{0pt}{0.1cm}\\[0.1cm] 
\hline 
$\meana{i}, \N{i}$ & $ \frac{\rewt{i} + 1}{\ndrawt{i} + 2}  ,\ndrawt{i} + 3 $ & $ \frac{\rewt{i}}{\ndrawt{i}} ,\ndrawt{i}$ \rule[0pt]{0pt}{0.4cm}\\[0.1cm] 
\hline
$\V{i}, \Deltam,  \Vtot, $ & $\frac{\meana{i} (1-\meana{i})}{\N{i}}, \meana{\maxa} - \meana{\mina}, \frac{\meana{\maxa} (1-\meana{\maxa})}{\N{\maxa}} + \frac{\meana{\mina} (1-\meana{\mina})}{\N{\mina}}$ & $\frac{\sigma^2}{\N{i}}, \meana{\maxa} -\meana{\mina}, \frac{\sigma^2}{\N{\maxa}} + \frac{\sigma^2}{\N{\mina}}$ \rule[0pt]{0pt}{0.4cm}\\[0.1cm] 
\hline
$\teq$  & $\meana{\maxa} + \sqrt{ 2 \Vmax [\N{\mina}  \kull(\meana{\mina},\meana{\maxa})  +  \frac{1}{2} \ln \frac{\N{\maxa}}{\N{\mina}} ]}$ & $\frac{\N{\maxa} \meana{\maxa} -\N{\mina} \meana{\mina}}{\N{\maxa}-\N{\mina}} + \sqrt{ \frac{4 \N{\maxa} \N{\mina}  (\meana{\maxa}-\meana{\mina})^2}{(\N{\maxa} - \N{\mina})^2}  + \frac{\sigma^2 \ln \frac{\N{\maxa}}{\N{\mina}} }{|\N{\maxa} -\N{\mina}|} } $ \rule[0pt]{0pt}{0.6cm}\\[0.2cm] 
\hline
$\ptail$ & $1- \BetaIncS{\teq}(\rewt{\mina} + 1, \ndrawt{\mina} -\rewt{\mina} +1) $& $\frac{1}{2} \erfc\left(  \frac{\sqrt{\N{\mina}} (\teq - \meana{\mina})}{\sqrt{2 \sigma^2}}\right)$ \rule[0pt]{0pt}{0.6cm}\\[0.3cm] 
\hline
$\Sc$ & \makecell{$ \frac{1}{4}\ln ( 2\pi \Vmax e^{1-2\Ac} ) +  \frac{1}{4}\ln ( 2\pi \Vmax e^{1+2\Ac} ) \erf\big[ \frac{\Delta}{\sqrt{2\Vtot}} \big] $\\ $- \frac{\Deltam \Vmax }{2 \sqrt{2 \pi }  \Vtot^{3/2}} e^{-\Deltam^2/2\Vtot}$ } &   \makecell{$ \frac{1}{4}\ln ( 2\pi \Vmax e^{1-2\Ac} ) +  \frac{1}{4}\ln ( 2\pi \Vmax e^{1+2\Ac} ) \erf\big[ \frac{\Delta}{\sqrt{2\Vtot}} \big] $\\ $- \frac{\Deltam \Vmax }{2 \sqrt{2 \pi }  \Vtot^{3/2}} e^{-\Deltam^2/2\Vtot}$ } \rule[0pt]{0pt}{0.8cm}\\[0.4cm] 
\hline 

$\Stail$ & $ (1- \BetaIncS{\teq, m}) [ \N{\mina} \kull(\meana{\mina}, \teq) + \frac{1}{2} \ln(2 \pi \Vmin )] $& $\frac{1}{4} \ln( 2\pi \Vmin e) \erfc \left(\frac{(\teq-\meana{\mina})}{\sqrt{2 \Vmin}} \right) + \frac{(\teq-\meana{\mina})}{2 \sqrt{2\pi \Vmin}} e^{ -\frac{-(\teq-\meana{\mina})^2 }{2 \Vmin}} $ \rule[0pt]{0pt}{0.7cm}\\[0.3cm] 
\hline
\end{tabular}
\caption{\label{table:maintable}
Analytic expressions for the terms of the approximate entropy Eq.~\eqref{theoryS} for Bernoulli (left) and Gaussian (right) reward distributions.
To derive closed-form analytical expression for the Bernoulli bandits, we applied a Laplace approximation. 
}
\end{table*}






\subsection{Gaussian approximation of the Beta posterior distribution}\label{Betaapprox}

For the Bernoulli bandits, we  approximate the Beta distributions by Gaussian distributions. 
To do so, we define $\meana{i}, \N{i}$ such that 
\begin{equation}\label{Siapprox1}
\mathbb{E}\left[\BetaDisLaw{\rew{i}}{\ndraw{i} -  \rew{i}}\right] = \frac{ \rew{i}+1}{\ndraw{i}+2} = \meana{i},
\end{equation}
and 
\begin{equation}\label{Siapprox2}
\begin{split}
\mathrm{Var}\left[\BetaDisLaw{\rew{i}}{\ndraw{i} -  \rew{i}} \right] &= \frac{ \rew{i}+1}{\ndraw{i}+2} \left(1 - \frac{ \ndraw{i} - \rew{i}+1}{\ndraw{i}+2} \right) \frac{1}{\ndraw{i}+3}\\ &= \frac{\meana{i} (1-\meana{i})}{\N{i}}.
\end{split}
\end{equation}
Thus, $\meana{i}$ and $\N{i}$ are respectively the mean and the number of draws that lead to a Gaussian approximation with the same two first moments as the true Beta distribution. 
(Note that for a Gaussian reward distribution, we have directly $\meana{i}=  \rew{i}/\ndraw{i}$ and  $\N{i}=\ndraw{i}$.) 

\subsection{ The partitioning approximation}\label{refteqapprox}
    
In this section we derive an approximation of the intersection point (defined above as $\teq$) where the distributions $\psupmaxmin$ and $\psupminmax$ intersect at their highest value (if more than one solution exists). 

We start with the case of Bernoulli bandits. The exact equation verified by the intersection point $\tp$ is
\begin{equation}\label{Sithetaeq1}
\begin{split}
&e^{- \N{\maxa} \kull(\meana{\maxa},\tp)} \int_0^{\tp} e^{- \N{\mina} \kull(\meana{\mina},\theta')} d\theta' = e^{- \N{\mina} \kull(\meana{\mina},\tp)} \int_0^{\tp} e^{- \N{\maxa} \kull(\meana{\maxa},\theta')} d\theta'.\\
\end{split}
\end{equation}
Taking the logarithm of Eq.~\eqref{Sithetaeq1} and normalizing the last term leads to
\begin{equation}\label{Sithetaeq2}
\begin{split}
&\N{\mina}  \kull(\meana{\mina},\tp) - \N{\maxa}\kull(\meana{\maxa},\tp)  +  \frac{1}{2}\ln \frac{\N{\maxa}}{\N{\mina}} + \ln \frac{\int_{0}^{\tp} \sqrt{\N{\mina}} e^{- \N{\mina} \kull(\meana{\mina},\theta')} d\theta')}{ \int_{0}^{\tp} \sqrt{\N{\maxa}} e^{- \N{\maxa} \kull(\meana{\maxa},\theta')} d\theta'} = 0.
\end{split}
\end{equation}

The distributions are uni-modal, and assuming that $(\meana{\maxa},\N{\maxa}) > (\meana{\mina},\N{\mina})$ and recalling that $\tp$ is the highest intersection solution, we approximate $\tp$ by neglecting the last term,
\begin{equation}\label{Sithetaeq3}
\begin{split}
  & \N{\mina}  \kull(\meana{\mina},\tp) - \N{\maxa}  \kull(\meana{\maxa},\tp)   + \frac{1}{2}\ln \frac{\N{\maxa}}{\N{\mina}}  \approx 0.
\end{split}
\end{equation}

In the long time limit $\N{\maxa} \gg \N{\mina}$ and $\tp$ will be in the vicinity of $\meana{\maxa}$ when the Gaussian expansion of the \kulleib divergence is relevant [in particular for $\N{\mina} \sim O(\ln\N{\maxa})$]. 
Thus, we approximate $\kull(\meana{\mina},\tp)$ by $\kull(\meana{\mina},\meana{\maxa})$ and expand $\kull(\meana{\maxa},\tp)$ to lowest order in $\tp$, which leads to the expression given in Table~\ref{table:maintable},
\begin{equation}\label{Sithetaeq4}
\begin{split}
&\teq =  \meana{\maxa} +\sqrt{ 2 \Vmax \left[ \N{\mina} \kull(\meana{\mina},\meana{\maxa}) + \frac{1}{2 }\ln \frac{\N{\maxa}}{\N{\mina}} \right]},
\end{split}
\end{equation}
where $ \Vmax = \meana{\maxa}(1- \meana{\maxa})/N_{\maxa}$, which verifies $\teq -\meana{\maxa} \sim o(\N{\maxa}^{-1/3})$, consistent with the Gaussian expansion of the \kulleib distance around $\meana{\maxa}$. 

We apply the same reasoning for Gaussian rewards, which leads to: 
\begin{equation}\label{Sithetaeq5}
\begin{split}
&\N{\mina} \frac{(\tp-\meana{\mina})^2}{2 \sigma^2} - \N{\maxa} \frac{(\tp-\meana{\maxa})^2}{2 \sigma^2}   +  \frac{1}{2} \ln \frac{\N{\maxa}}{\N{\mina}}  \approx 0.
\end{split}
\end{equation}
Solving for $\tp$ leads to: 
\begin{equation}\label{Sithetaeq6}
\begin{split}
&\teq = \frac{(\N{\maxa} \meana{\maxa} -\N{\mina} \meana{\mina})}{\N{\maxa}-\N{\mina}} + \frac{2}{|\N{\maxa}-\N{\mina}|}  \times  \sqrt{ \N{\maxa}\N{\mina} ( \meana{\maxa}-\meana{\mina})^2  + \sigma^2 (\N{\maxa}- \N{\mina} )\ln \left( \frac{\N{\maxa}}{\N{\mina}} \right) } .
\end{split}
\end{equation}

Note that the expressions, Eq.~\eqref{Sithetaeq4} and Eq.~\eqref{Sithetaeq6}, rely on the assumption  that $(\meana{\maxa},\N{\maxa}) > (\meana{\mina},\N{\mina})$. 
For $\N{\maxa} \leq \N{\mina}$, the contributions from $\parmin$ and $\parmax$ do not intersect and $\teq=\bsup$ (i.e., $\teq =1$ and $\teq=+\infty$ for Bernoulli and Gaussian rewards, respectively), which means that the contribution from the tail is zero.


\subsection{Closed-form expressions for the main mode's contribution}\label{SiScgaussian}

\subsubsection{Gaussian posterior distributions}

Here, we derive the $\Sc$ term given in Table~\ref{table:maintable}  for Gaussian posterior distributions. 
Inserting the Gaussian form of the posterior into Eq.~\eqref{Sbodyexp} gives:
\begin{equation}\label{SiSb1}
\begin{split}
\Sc = &-\int_{-\infty}^{+\infty} \frac{e^{- \renormmax^2} }{\sqrt{2\pi \Vmax}}   \frac{1}{2}\big[ 1 + \erf (\renormmin ) \big] \left(-\frac{1}{2}\ln( 2 \pi \Vmax) - \renormmax^2 \right) \\ &-\int_{-\infty}^{+\infty} \Ac   \frac{e^{- \renormmin^2} }{\sqrt{2\pi \Vmin}} \frac{1}{2}\big[ 1 + \erf (\renormmax ) \big]  d\theta,
\end{split}
\end{equation}
where $V_{i}$ is the distribution's variance, $\renormmax =  (\mean - \meana{\maxa})/\sqrt{2 \Vmax}$, and $\renormmin =  (\mean - \meana{\mina})/\sqrt{2 \Vmin}$. 
We integrate the constant part of the first term by use of the following identity~\cite{ng_table_1969}:
\begin{equation}\label{Sierfint}
\begin{split}
\int_{-\infty}^\infty \frac{1}{2} & \bigg[1+\erf \left(\frac{\theta-\theta_1}{\sqrt{2V_1}} \right) \bigg] \frac{ e^{-\frac{(\theta-\theta_2)^2}{2 V_2}}}{\sqrt{2 \pi V_2}} =  \frac{1}{2} \left[1  +  \erf \left(\frac{\theta_2-\theta_1}{\sqrt{2} \sqrt{V_2 + V_1}} \right) \right],
\end{split}
\end{equation}
which leads to 
\begin{equation}\label{SiSb2}
\begin{split}
  &\int_{-\infty}^\infty \frac{1}{2} \frac{e^{-\renormmax^2} }{\sqrt{2\pi \Vmax}} \left[1 + \erf(\renormmin) \right]  \frac{1}{2} \ln \left(2\pi \Vmax \right)  d\theta = \frac{1}{4}\ln \left(2\pi \Vmax \right)  \left[ 1 + \erf \left(\frac{\meana{\maxa}-\meana{\mina}}{\sqrt{2(\Vmax + \Vmin)}} \right)  \right].
\end{split}
\end{equation}

Next, we integrate by parts the second part of the first term to obtain:
\begin{equation}\label{SiSb3}
\begin{split}
 \int_{-\infty}^\infty  \renormmax^2 \frac{1}{2}\left[1 + \erf(\renormmin)\right] \frac{e^{-\renormmax^2}}{\sqrt{2\pi\Vmax}} &= \int_{-\infty}^\infty \frac{1}{4}  \frac{e^{-\renormmax^2}}{\sqrt{2\pi\Vmax}} \left[1 + \erf(\renormmin)\right]
+ \int_{-\infty}^\infty (\mean-\meana{\maxa})  \frac{1}{2}\frac{e^{-\renormmax^2} }{\sqrt{2\pi \Vmax}} \frac{ e^{-\renormmin^2}}{\sqrt{2\pi \Vmin}} \\
     &=  \frac{1}{4} \left[1  +  \erf \left(\frac{\meana{\maxa} - \meana{\mina}}{\sqrt{2(\Vmax + \Vmin)}} \right) \right]
     +\frac{(\meana{\mina} -\meana{\maxa}) \Vmax}{2 \sqrt{2 \pi}(\Vmax +\Vmin)^{3/2}} e^{-\frac{(\meana{\maxa} - \meana{\mina})^2}{2(\Vmax + \Vmin)}}, 
\end{split}
\end{equation}
where we also employed the identity of Eq.~\eqref{Sierfint}. 

Finally, the last term is also integrated using Eq.~\eqref{Sierfint}, giving:
\begin{equation}\label{SiSb4}
 -\Ac \int_{-\infty}^\infty \frac{1}{2} \left[1+ \erf(\renormmax) \right] \frac{ e^{-\renormmin^2}}{\sqrt{2 \pi \Vmin}}  =  -\frac{\Ac}{2} \left[1  +  \erf \left(\frac{\meana{\mina}-\meana{\maxa}}{ \sqrt{2(\Vmax +\Vmin})} \right) \right].
\end{equation}

Combining Eq.~\eqref{SiSb2}, Eq.~\eqref{SiSb3} and Eq.~\eqref{SiSb4} leads to the expression given in Table~\ref{table:maintable}. 

\subsubsection{Bernoulli posterior distributions}

Reminding \ref{SiScgaussian}, the analytic derivation of $\Sc$ made for Gaussian reward [Eq.~\eqref{SiSb2}, Eq.\eqref{SiSb3} and Eq.~\eqref{SiSb4}] can be extended to Bernoulli reward with $\meana{i}$ and $\N{i}$ thus obtained.

\subsection{Closed-form expressions for the tail contribution}\label{Sitail}

We conclude our approach by considering the tail contribution to the approximate entropy.

\subsubsection{Gaussian posterior distribution}

We first consider the Gaussian reward case for which the contribution from the tail can be derived exactly,
\begin{equation}\label{SiSb5}
\begin{split}
\Stail &= \int_{\theta_{eq}}^{\infty} \frac{e^{-\renormmin^2}}{\sqrt{2\pi \Vmin}} \bigg[ \frac{1}{2}\ln(2\pi \Vmin) + \renormmin^2 \bigg]\\ 
&= \frac{1}{4}\ln(2\pi \Vmin e) \erfc \left( \frac{\teq-\meana{\mina}}{\sqrt{2\Vmin}} \right) + \frac{\teq-\meana{\mina}}{2\sqrt{2\pi \Vmin}} e^{ -\frac{(\teq-\meana{\mina})^2}{2 \Vmin}}.
\end{split}
\end{equation}

\subsubsection{Bernoulli posterior distribution}

We now focus on the Bernoulli case for which $\Stail$ requires a second approximation in order to get rid of the numerical integration. We have:
\begin{equation}\label{SiSb6}
\begin{split}
\Stail &\approx  -\ptail \ln(p_{\meana{\mina}}(\teq)) \\
&\approx  \left(1- \BetaIncS{\teq, m}\right) \left[ \N{\mina} \kull(\meana{\mina}, \teq) + \frac{1}{2} \ln(2 \pi \Vmin ) \right],
\end{split}
\end{equation}
where $\BetaIncS{\teq, m}$ is the normalized incomplete beta function evaluated at $\teq$ with parameters $\rew{\mina} + 1$ and $\ndraw{\mina} - \rew{\mina} + 1$. Of note, we have bounded $\ln(p_{\meana{\mina}}(\theta))$ by its value at $\teq$ and included only the leading order at large times. 
We remark that Eq.~\eqref{SiSb6} is one possible solution, but others would work as well as long as their leading order is given by $\ptail$. 

Applying Eq.~\eqref{greedygradient} to the obtained analytic expression Eq.~\eqref{theoryS}, leads to the \algoname algorithm.

\section{AIM algorithms}\label{SiAIMimp}

Here, we summarize the algorithm procedures introduced in the main text. 

\subsection{Two-armed Gaussian bandit}\label{gaussianalgo}

\begin{enumerate}

\item Draw each arm once, update $\rewt{i}, \ndrawt{i}$ and their associated $\meana{i}, \N{i}$ according to Table~\ref{table:maintable}.

\item  For $t > 2$, sort the arms according to $\meana{i}(t)$ to get $(\meana{\mina}(t), \N{\mina}(t), \ndrawt{\mina}, \rewt{\mina})$ and $(\meana{\maxa}(t), \N{\maxa}(t), \ndrawt{\maxa}, \rewt{\maxa})$ couples.

\begin{itemize}[leftmargin=0.1in]
\item If $\meana{\mina}(t) =  \meana{\maxa}(t)$ then choose the arm which has been currently drawn the least. If $\N{\mina}(t) =  \N{\maxa}(t)$ choose randomly. 
\item Otherwise, if $ \N{\mina}(t) \geq \N{\maxa}(t) $ then draw $\arma{\maxa}$.
\item Otherwise, evaluate $\teq$ according to Table~\ref{table:maintable}. Then, evaluate the absolute value of the gradient  of $\Sapp( \rewt{i}, \ndrawt{i}, \rewt{j}, \ndrawt{j}, \teq)$ along each arm according to:  
%
%
\begin{equation}\label{gradientgauss}
\begin{split}
\hspace{1cm} \Delta_i \Sapp = &\bigg| \frac{1}{2} \Sapp( \rewt{i} + \meana{i}(t) +  \alpha \sigma, \ndrawt{i} + 1, ..) + \frac{1}{2} \Sapp(\rewt{i}  + \meana{i}(t) - \alpha \sigma, \ndrawt{i} + 1, ..) - \Sapp(\rewt{i}, \ndrawt{i},..) \bigg|,\\
\end{split}
\end{equation}
where the dots refer to constant variables ($\rewt{j}$, $\ndrawt{j}$, $\teq$), $\alpha=1$, and $\Sapp$ given by Eq.~(\ref{theoryS}) with Table~\ref{table:maintable}. Next, draw the arm with the highest gradient.
\end{itemize}
\item Update $\rew{i}(t+1), \ndraw{i}(t+1)$  according to the reward returned by the chosen arm.
\end{enumerate}

Let us draw some additional observations. First, we stress that if $\N{\maxa} \leq \N{\mina}$, the current best arm, $\arm_{\maxa}$, is automatically drawn. It is because the current best arm should always be played in this  case (since the information about it is lesser than the current worst arm, $\arm_{\mina}$). 
Next, $\Sapp$ also requires to sort its input values. Thus,  $\meana{\maxa}(t), \N{\maxa}(t), \meana{\mina}(t)$ and $\N{\mina}(t)$ used in each $\Sapp$ evaluation are different from the ones used in the sorting step $2$. However, $\teq$ is shared among all $\Sapp$ evaluations. This avoids adding perturbations induced by the cutoff of the tail. Finally, one should note that Eq.~\ref{gradientgauss} is a particular way to evaluate the gradient, but other approaches are possible (i.e., by modifying $\alpha$ or computing the higher derivative orders).

\subsection{Two-armed Bernoulli bandit}\label{entropyalgo}

For the Bernoulli bandit, most of the procedure is identical to the Gaussian case described above. One simply has to replace the expressions for the case of Gaussian rewards by those corresponding to Bernoulli rewards given in Table~\ref{table:maintable}. 
The only difference in the procedure regards the gradient evaluation [Eq.~\eqref{gradientgauss}, which is replaced by:
\begin{equation}\label{gradientber}
\begin{split}
\Delta_i \Sapp = & \bigg|\frac{\rewt{i}}{\ndrawt{i}}
\Sapp( \rewt{i} +1, \ndrawt{i} + 1,..) + \frac{\ndrawt{i} -\rewt{i}}{\ndrawt{i}} \Sapp(\rewt{i},  \ndrawt{i} + 1,..)  - \Sapp( \rewt{i}, \ndrawt{i},..) \bigg|,\\
\end{split}
\end{equation}
%
%
where, as above, the two dots refer to constant variables $\rewt{j}$, $\ndrawt{j}$ and $\teq$.

\subsection{$\mathrm{K>2}$ armed Gaussian bandit}\label{multigaussianalgo}

To further assess the efficiency of the \algoname algorithm, we address the multi-armed case (with a number of arms $K>2$). 
We notice that Eq.~\eqref{theoryS} is asymmetric in $(\arma{\max},\arma{i})$, which suggests that all but $\arma{\max}$ could be decoupled in the general entropy expression by neglecting the correlations between subdominant arms. 
In practice, we thus propose to evaluate the $K-1$ gradients between each subdominant arm and $\arma{\max}$ by the use of Eq.~\eqref{theoryS} with each subdominant arm in place of $\arm_{\mina}$. The dominant arm $\arma{\max}$ is pulled if all the gradient evaluations favor $\arma{\max}$, and the subdominant arm with the highest absolute gradient is chosen if at least one gradient favors a subdominant arm. Thus, the \algoname implementation for $K>2$ reads: 

\begin{enumerate}

\item Draw each arm once, and update $\rewt{i}, \ndrawt{i}$ and their associated $\meana{i}, \N{i}$ according to Table~\ref{table:maintable}.
 \item At each step $t > K$, determine the arm with the best empirical mean reward such that:
 \begin{itemize}
 \item $\meana{\maxa} > \meana{i} \:\: \forall i \neq \maxa$,
 \item or  $i_{\maxa} = \underset{ \{ i : \meana{i} = \meana{\maxa} \} }{ \mathrm{argmax}( \N{i} )  } $ if the dominant arm is not unique. If the maximal $\N{i}$ is not unique, $i_{\maxa}$ is drawn randomly among $\{i : \meana{i} = \meana{\maxa} \land \N{i} = \N{\maxa}\}$.
 \end{itemize}

 \item Compare each arm $i$ to $\arma{\maxa}$ by computing the two following gradients:
%
%
\begin{equation}
\begin{split}
\Delta_i \Sapp = \bigg| \frac{1}{2} &\Sapp( \rewt{i} + \meana{i}(t) +  \alpha \sigma, \ndrawt{i} + 1, ..) + \frac{1}{2}\Sapp( \rewt{i} + \meana{i}(t) -  \alpha \sigma, \ndrawt{i} + 1, ..)  - \Sapp(\rewt{i},\ndrawt{i},..)  \bigg|,
\end{split}
\end{equation}
where the two dots refer to constant variables ($\rewt{\maxa}$, $\ndrawt{\maxa}$, $\teq$), and 
\begin{equation}
\begin{split}
\Delta_{i, \maxa} \Sapp =  \bigg| \frac{1}{2} \Sapp( \rewt{\maxa} + \meana{\maxa}(t) +  \alpha \sigma, \ndrawt{\maxa} + 1,.. ) + \frac{1}{2} \Sapp(  \rewt{\maxa} + \meana{\maxa}(t) - \alpha \sigma, \ndrawt{\maxa} + 1,.. ) - \Sapp( \rewt{\maxa}, \ndrawt{\maxa},..) \bigg|,\\
\end{split}
\end{equation}
where the two dots refer to constant variables $\rewt{i}$, $ \ndrawt{i}$ and $\teq$, and  $\teq$ is computed following Table~\ref{table:maintable} with $\rewt{i}, \ndrawt{i}, \rewt{\maxa}, \ndrawt{\maxa}$.
 
\item Then, select the arm $\Armt$ such that:
\begin{itemize}
\item $\Armt = \arma{\maxa} $ if $\Delta_i \Sapp < \Delta_{i, \maxa} \Sapp, \forall i \neq \maxa$,
\item or $\Armt = \underset{\Delta_i \Sapp  - \Delta_{i, \maxa} \Sapp\geq 0}{ \{ \mathrm{argmax}(\Delta_{i} \Sapp - \Delta_{i, \maxa} \Sapp) \}}$ elsewise. If there is more than one solution $\Armt$ is drawn randomly among them.
\end{itemize}

\item Update  $\rew{i}(t+1), \ndraw{i}(t+1)$ according to the reward returned by the chosen arm.
\end{enumerate}

\subsection{$\mathrm{K>2}$ armed Bernoulli bandit}\label{multiberalgo}

For the multi-armed Bernoulli bandit, most of the procedure is identical to the multi-armed Gaussian algorithm described above. 
As for the two-armed case, the procedure is implemented by replacing the expressions by those for Bernoulli rewards given in Table~\ref{table:maintable}. 
The expressions for the gradients are replaced by:
 %
%
\begin{equation}\label{Bernoullimulitgrad}
\begin{split}
\Delta_i \Sapp = \bigg| \frac{\rewt{i}}{\ndrawt{i}} &\Sapp( \rewt{i} + 1, \ndrawt{i} + 1, ..) + \frac{\ndrawt{i} -\rewt{i}}{\ndrawt{i}} \Sapp( \rewt{i}, \ndrawt{i} + 1, ..)  - \Sapp(\rewt{i},\ndrawt{i},..)  \bigg|,
\end{split}
\end{equation}
where the two dots refer to constant variables ($\rewt{\maxa}$, $\ndrawt{\maxa}$, $\teq$), and 
\begin{equation}\label{Bernoullimulitgrad2}
\begin{split}
\Delta_{i, \maxa} \Sapp =  \bigg|\frac{\rewt{\maxa}}{\ndrawt{\maxa}} \Sapp( \rewt{\maxa} +1, \ndrawt{\maxa} + 1,..) + \frac{\ndrawt{\maxa} -\rewt{\maxa}}{\ndrawt{\maxa}} \Sapp(  \rewt{\maxa}, \ndrawt{\maxa} + 1,..) - \Sapp( \rewt{\maxa}, \ndrawt{\maxa},..) \bigg|,\\
\end{split}
\end{equation}
where the two dots refer to constant variables $\rewt{i}$, $ \ndrawt{i}$ and $\teq$.

\section{Other state-of-the-art bandit algorithms}\label{Simethodsalgo}

Here, we briefly review several baseline algorithms which provide a benchmark of our gradient method.

\subsection{Epsilon-n-Greedy}

This method is a variation of the \egreedy strategy, and is one of the most widely used bandit algorithms due to its undeniable simplicity \cite{sutton_reinforcement_1998}. 
The \egreedy strategy selects either a random arm with a probability $\epsilon$ or the current dominant arm otherwise. The \engreedy strategy is a generalized form of this approach where the parameter $\epsilon$ is a time-dependent function $\epsilon(t) = \mathrm{min} \{ 1, c(\marm_1,\marm_2) K/(d^2 t) \} $. The constant $c$ is a hyperparameter of the method, which needs to be tuned for optimal performance. Here, we used $c=10$ tuned for Bernoulli uniform priors and $c=30$ for Gaussian uniform priors. Let us stress that \engreedy relies on a priori knowledge of the distribution of $\{\marm_1,\marm_2\}$ in order to be effective.  

\subsection{UCB-Tuned}

This method belong to the class of upper confidence bound (UCB) algorithms which pull the arm maximising a proxy function generally defined as $F_i = \meana{i} + R_i$ where $R_i$ bounds the regret to logarithmic growth. 
For UCB-tuned, $R_i$is given by:
\begin{equation}\label{UCBTeq}
\begin{split}
R_i &= c(\marmi{1},\marmi{2}) \sqrt{ \frac{\ln(t)}{\ndrawt{i}} \mathrm{min} \left( \frac{1}{4}, s_i(t) \right) },\: \: s_i(t) = \hat{\sigma_i}^2 + \sqrt{ \frac{2\ln(t)}{\ndrawt{i}}},
\end{split}
\end{equation}

where $\hat{\sigma_i}^2$ is the reward variance and $c$ a hyperparameter.
Here, we rely on the optimised version of $F_i$ proposed in~\cite{pilarski_optimal_2021} for Bernoulli reward: 
\begin{align}\label{UCBTeq2}
F_i &= \frac{\rewt{i} + 1}{\ndrawt{i} + 2} + c(\marmi{1},\marmi{2}) \sqrt{ \frac{\ln(t+2K)}{\ndrawt{i}+2} \mathrm{min} \big( d, s_i(t) \big) }, \: s_i(t) = \frac{(\rewt{i} + 1)(\ndrawt{i} - \rewt{i} + 1)}{(\ndrawt{i}+2)^2 (\ndrawt{i}+
3)} + \sqrt{ \frac{2\ln(t+2K)}{\ndrawt{i}+2}},
\end{align}
with $c=0.73$ and $d=0.19$. For Gaussian rewards we adapt \eqref{UCBTeq} with $c=2.1$ and  $\hat{\sigma_i}^2 = \frac{\sigma^2}{\ndrawt{i}}$, although this is not necessarily optimal.

\subsection{KL-UCB}

This method is an another upper confidence bound (UCB) variant which has been especially designed for bounded reward, and in particular for Bernoulli distributed rewards where it reaches the Lai and Robbins bound~\cite{garivier_kl-ucb_2011}. For KL-UCB, $F_i$ reads: 

\begin{equation}\label{KLUCB}
\begin{split}
F_i = \mathrm{\max} \Bigl\{ &\theta \in \Theta: \N{i} \kull \left( \frac{\rewt{i}}{\ndrawt{i}}, \theta \right) \leq \ln(t) + c(\marmi{1},\marmi{2})\ln(\ln( t)) \Bigl\},
\end{split}
\end{equation}

where $\Theta$ denotes the definition interval of the posterior distribution. By testing various $c$ values, we end up with  $c(\marmi{1},\marmi{2}) = 0.00001$. 

\subsection{Thompson sampling}

At each step, Thompson sampling~\cite{thompson_likelihood_1933,kaufmann_thompson_2012} stochastically selects an arm based on the posterior probability that it maximizes the expected reward. 
In practice, after drawing $K$ random values according to each arms' posterior distribution, it picks the arm with the largest value:
\begin{equation}\label{Thomposn}
\Armt = \underset{i=1..K}{\mathrm{argmax}} \bigg( \BetaDisLaw{\rew{i}}{\ndraw{i} -  \rew{i}} \bigg).
\end{equation}

\subsection{Infomax}

At each step, Infomax relies on a greedy entropy minimization to decide the arm to be played. Here we adapt the Infomax~\cite{reddy_infomax_2016, vergassola_infotaxis_2007} algorithm by replicating the steps detailed in Supplementary Section \ref{entropyalgo} but replacing $\Sapp$ by a numerical integration of $\Smax$ [Eq.~\eqref{Smax}].
The regret performance of the Infomax algorithm thus obtained is compared with \algoname and Thompson methods in Fig.~\ref{fig:2sup}.

\begin{figure}[H]
\centering
\includegraphics[scale=0.6]{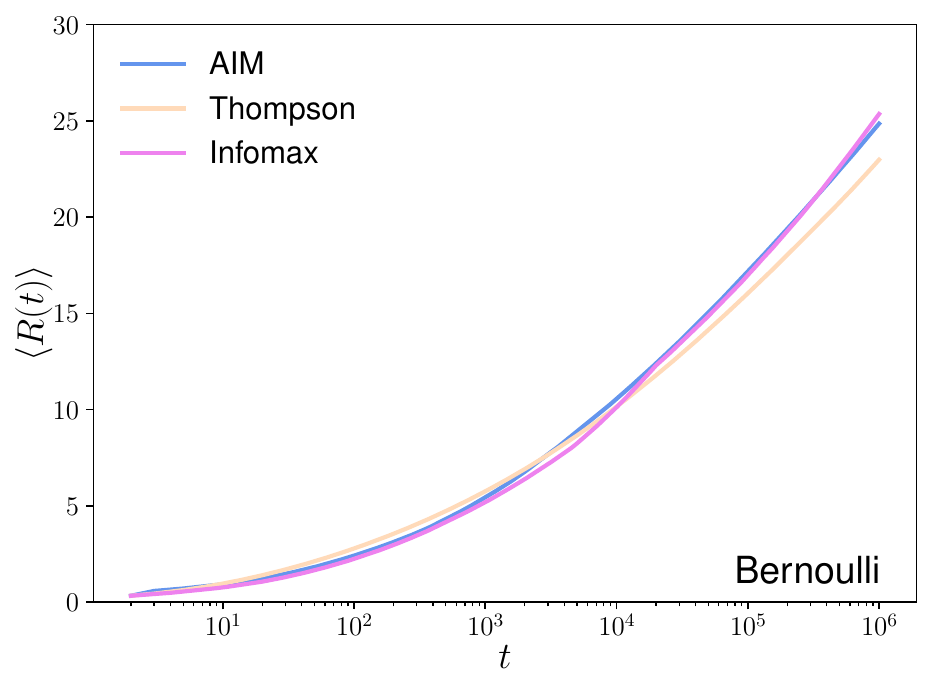}
\caption{ Comparison of AIM and Thompson sampling with the Infomax algorithm based on the exact entropy. Mean regret as function of time for 2-armed Bernoulli bandits with parameters drawn uniformly in $]0,1[$. In light blue \algoname, in orange Thompson, and in pink our Infomax gradient implementation (see \ref{Simethodsalgo} and \ref{SiAIMimp}) with the exact entropy obtained by numerical integration. The algorithms relying on the exact entropy or its analytic approximation show very similar performance. }
\label{fig:2sup}
\end{figure}

\section{Asymptotic optimality of approximate information maximisation}\label{refasymptotepart}

Here, we provide details on the asymptotic behaviour of the entropy at fixed $\meana{\maxa}$, $\meana{\mina}$, and $t\gg1$.  Recall that the derivative of Eq.~\eqref{theoryS}, from which we seek to derive the asymptotic behavior of $\N{\mina}$ by minimizing $\Sapp$, is given by:
\begin{equation}\label{SigradientSNmin}
\begin{split}
\frac{\partial \Sapp}{\partial \N{\mina} } &=   (1-\ptail) \frac{\partial \Sc}{\partial \N{\mina} } + \frac{\partial \Stail}{\partial \N{\mina} }  +  (-\Sc  + \ln(1-\ptail) + 1) \frac{\partial \ptail}{\partial \N{\mina} }   .
\end{split}
\end{equation}

We will focus on Bernoulli reward distributions, and thus the terms of $\Sapp$ given by Table~\ref{table:maintable}.
Since the different terms share common derivatives, we first provide the $\teq$ derivative. To leading order it is equal to:
\begin{equation}\label{SigradientSNmin1}
\begin{split}
 \frac{\partial \teq}{\partial  \N{\mina}} &= \frac{\meana{\maxa}(1-\meana{\maxa})}{\teq-\meana{\maxa}}\frac{t(2 \N{\mina} \kull(\meana{\mina}, \meana{\maxa}) -1) + \N{\mina} \ln(t/\N{\mina}  -1)  }{2(t- \N{\mina})^2 \N{\mina}}\\
 &=  O \left( \frac{\ln(t)}{\sqrt{t \N{\mina}}}, \sqrt{ \frac{\ln(t)}{\sqrt{t}}}  \right).
 \end{split}
\end{equation}

Next, we focus on the norm and main mode terms, $\ptail$ and $\Sc$, respectively. 
Since $\Vmax, \Vtot \ll \Vmin$, all the terms 
that depend exponentially on $\Vmax$ or $\Vtot$  are negligible to leading order. 
We thus obtain 
%
%
%
\begin{equation}\label{SigradientSNmin3}
 \frac{\partial \Sc}{\partial  \N{\mina}} \sim \frac{1}{2(t-  \N{\mina})}  + O(\exp(-C_0 t)),
\end{equation}
with $C_0 >0$. 

Next, we consider the tail terms of which we propose to rewrite the regularized incomplete beta distribution as 
\begin{equation}\label{SigradientSNmin3bis}
\begin{split}
 \ptail &= 1- \BetaIncS{\teq,m}\\
 &= C(\mmin,\nmin)\sqrt{\nmin} \int_{\teq}^1 e^{-\nmin \kull (\mmin,\theta)} d\theta \\
                   &= C(\mmin,\nmin)\sqrt{\nmin} e^{-\nmin \kull (\mmin,\teq)} \int_{\teq}^1 e^{-\nmin \left[\mmin \ln( \frac{\teq}{\theta}) + (1-\mmin) \ln( \frac{1-\teq}{1-\theta}) \right]  } d\theta,
 \end{split}
\end{equation}
where $\mmin = \frac{\rew{\mina}}{\ndraw{\mina}}$, $\nmin = \ndraw{\mina}$ and $C(\mmin,\nmin) \sim [2 \pi \mmin (1-\mmin)]^{-\frac{1}{2}} + O(\nmin^{-\frac{1}{2}})$ by convergence of the Beta distribution to its Gaussian counterpart. Next, we partition the integral, which denotes $I$, at a cutoff $\mu_c = \nmin (\meana{c} - \teq)$, leading to:
\begin{equation}\label{SigradientSNmin4}
\begin{split}
&I = \int_{\mean{c}}^1 e^{-\nmin \left[ \mmin \ln( \frac{\teq}{\theta}) + (1-\mmin) \ln( \frac{1-\teq}{1-\theta}) \right]  } d\theta + \int_{0}^{\mu_c} \frac{e^{\nmin \left[ \mmin \ln(1 + \frac{\mu}{\teq \nmin}) + (1-\mmin) \ln(  1 -\frac{\mu}{\nmin(1-\teq)}) \right]  }}{\nmin}d\mu. \\
 \end{split}
\end{equation}
Taking $\mu_c \sim A \sqrt{\nmin}$, we obtain a well-defined expansion of the second integral, which leads to: 
\begin{equation}\label{SigradientSNmin5}
\begin{split}
I &= C e^{\nmin \left[ \mmin \ln( 1 + \frac{A}{ \teq \sqrt{\nmin}}) + (1-\mmin) \ln( 1- \frac{A}{(1-\teq) \sqrt{\nmin} }) \right] }  + \int_{0}^{ A \sqrt{\nmin}} \frac{e^{\mu (\frac{\mmin}{\teq} - \frac{1-\mmin}{1-\teq}) }}{\nmin}  \bigg[ 1 + A_{\mina} \frac{\mu^2}{ \nmin} + O(\frac{\mu^3}{\nmin^2}) \bigg] d\mu \\
&= \frac{\teq (1-\teq)}{\N{\mina} (\teq-\meana{\mina})} + O(\N{\mina}^{-2}).
 \end{split}
\end{equation}
where the difference between $\meana{\mina}$ and $\mmin$ is of the order $O(\N{\mina}^{-1})$ and are then included in the second term.

%
By the use of Eq.~\eqref{SigradientSNmin5}, we find
\begin{equation}\label{SigradientSNmin7second}
\begin{split}
\ptail &\sim C_2 \frac{\teq (1-\teq)}{ \sqrt{\N{\mina}} (\teq-\meana{\mina})}  e^{- \N{\mina}\kull ( \meana{\mina},\teq)} \left[ 1 + O(\N{\mina}^{-{1}}) \right],
 \end{split}
\end{equation}
to leading order. 
Using this gives us the dominant order of the variation of $\ptail$:
\begin{equation}\label{SigradientSNmin6}
\begin{split}
\frac{ \partial \ptail}{\partial  \N{\mina}} &= -\frac{C_2}{2 \N{\mina}^{\frac{3}{2}}}  \frac{\teq (1-\teq)}{(\teq-\meana{\mina})}  e^{- \N{\mina}\kull ( \meana{\mina},\teq)} 
 - C_2\frac{\partial \teq}{\partial \N{\mina} } \left[1 + \frac{\meana{\mina}(1-\meana{\mina})}{(\meana{\mina} - \teq)^2} \right] \frac{e^{- \N{\mina}\kull ( \meana{\mina},\teq)}}{\sqrt{\N{\mina}}}  \\
  &-C_2 \frac{\teq (1-\teq)}{ \sqrt{\N{\mina}} (\teq-\meana{\mina})} e^{- \N{\mina}\kull ( \meana{\mina},\teq)}  \bigg[  \kull (\meana{\mina},\teq) +  \N{\mina}\frac{\partial \teq}{\partial \N{\mina} } \left(  \frac{1-\meana{\mina}}{1- \teq} - \frac{\meana{\mina}}{\teq}\right)  \bigg] \\
  &+\frac{\partial C_2}{\partial  \N{\mina}}  \frac{\teq (1-\teq)}{\sqrt{\N{\mina}}(\teq-\meana{\mina})}  e^{- \N{\mina}\kull ( \meana{\mina},\teq)} + O \left( e^{- \N{\mina}\kull ( \meana{\mina},\teq)} \N{\mina}^{-2} \right), \\
 \end{split}
\end{equation}
where $C_2 = C  e^{\N{\mina} \kull (\meana{\mina},\teq)}  e^{-\nmin \kull (\mmin,\teq)} $ accounts for the change of variable between $(\mmin,\nmin)$ and $(\meana{\mina},\N{\mina})$. The derivative of $C_2$ is to leading order:
\begin{equation}\label{SigradientSNmin7}
 \frac{\partial C_2}{\partial  \N{\mina}} \sim C_2 O \left(\frac{1}{\N{\mina}^{3/2}}\right).
 \end{equation}
Combining Eq.~\eqref{SigradientSNmin7} with Eq.~\eqref{SigradientSNmin6} yields the leading order of the derivative of $\ptail$:
\begin{equation}\label{SigradientSNmin7bis}
\begin{split}
\frac{ \partial \ptail}{\partial  \N{\mina}} &\sim  - C_2 \kull (\mmin,\teq)   \frac{\teq (1-\teq)}{ \sqrt{\N{\mina}} (\teq-\meana{\mina})}  e^{- \N{\mina}\kull ( \meana{\mina},\teq)} \left(1+ O \left(\frac{1}{\N{\mina}} \right)\right).
 \end{split}
\end{equation}

We finally expand $\Stail$ to leading order, which yields:
\begin{equation}\label{SigradientSNmin8}
\begin{split}
\frac{ \partial \Stail}{\partial  \N{\mina}}&= \frac{ \partial \ptail }{\partial  \N{\mina}} [ \N{\mina} \kull(\meana{\mina}, \teq) + \frac{1}{2} \ln(2 \pi \Vmin )] + \ptail \bigg[ \N{\mina}\frac{\partial \teq}{\partial \N{\mina} } \left(  \frac{1-\meana{\mina}}{1- \teq} - \frac{\meana{\mina}}{\teq}\right) + \kull(\meana{\mina}, \teq)  - \frac{1}{2 \N{\mina}}  \bigg]\\
&= - C_2 \kull (\mmin,\teq)^2   \sqrt{\N{\mina}}  \frac{\teq (1-\teq)}{(\teq-\meana{\mina})}  e^{- \N{\mina}\kull ( \meana{\mina},\teq)} \left(1+ O\left(\frac{1}{\N{\mina}} \right)\right).
\end{split}
\end{equation}
%
Inserting the leading order terms of Eqs. \eqref{SigradientSNmin3}, \eqref{SigradientSNmin6}, and \eqref{SigradientSNmin8} into Eq. \eqref{SigradientSNmin} leads to:
\begin{equation}\label{SigradientSNmin9}
\begin{split}
 \frac{\partial \Sapp}{\partial \N{\mina} }
 &= \frac{1}{2(t-  \N{\mina})} -  C_2  \sqrt{\N{\mina}}  \kull (\mmin,\teq)^2  \frac{\teq (1-\teq)}{  (\teq-\meana{\mina})}  e^{- \N{\mina}\kull ( \meana{\mina},\teq)} \left(1+ O \left(\frac{1}{\N{\mina}} \right)\right)  \\ \hspace{2cm} & - \frac{1}{2}\ln(t) C_2 \kull (\mmin,\teq)   \frac{\teq (1-\teq)}{ \sqrt{\N{\mina}} (\teq-\meana{\mina})}  e^{- \N{\mina}\kull ( \meana{\mina},\teq)} \left(1+ O \left(\frac{1}{\N{\mina}} \right)\right).
 \end{split}
\end{equation}

Finally, setting $\frac{\partial \Sapp}{\partial \N{\mina} } =0$ leads to: 
\begin{equation}\label{SigradientSNmin9bis}
\begin{split}
  \frac{1}{t} \sim 2 C_2  \sqrt{\N{\mina}}  \kull (\mmin,\teq)  \frac{\teq (1-\teq)}{  (\teq-\meana{\mina})}  e^{- \N{\mina}\kull ( \meana{\mina},\teq)} \left(   1 + \frac{\ln(t)}{2  \kull (\mmin,\teq)\N{\mina}}  \right),
 \end{split}
\end{equation}
which, by taking the logarithm and noting that $\teq \to \meana{\maxa}$, leads to:
\begin{equation}\label{SigradientSNmin10}
\begin{split}
\ln(t) \sim  \N{\mina}\kull ( \meana{\mina},\meana{\maxa}) - \ln \bigg[ 2 C_2  \sqrt{\N{\mina}}  \kull (\mmin,\meana{\maxa})  \frac{\meana{\maxa} (1-\meana{\maxa})}{  (\meana{\maxa}-\meana{\mina})}  \left(   1 + \frac{\ln(t)}{2  \kull (\mmin,\meana{\maxa})\N{\mina}} \right) \bigg].
 \end{split}
\end{equation}

Hence, we obtain the Lai and Robbins relation for the \algoname algorithm: 
\begin{equation}\label{SigradientSNmin10bis}
\begin{split}
\N{\mina} \sim \frac{\ln(t)}{\kull ( \meana{\mina},\meana{\maxa})} +  o(\ln(t)).\\
 \end{split}
\end{equation}

\section{Tuned approximate information maximization}\label{SItuned}


\begin{figure*}
\includegraphics[scale=0.7]{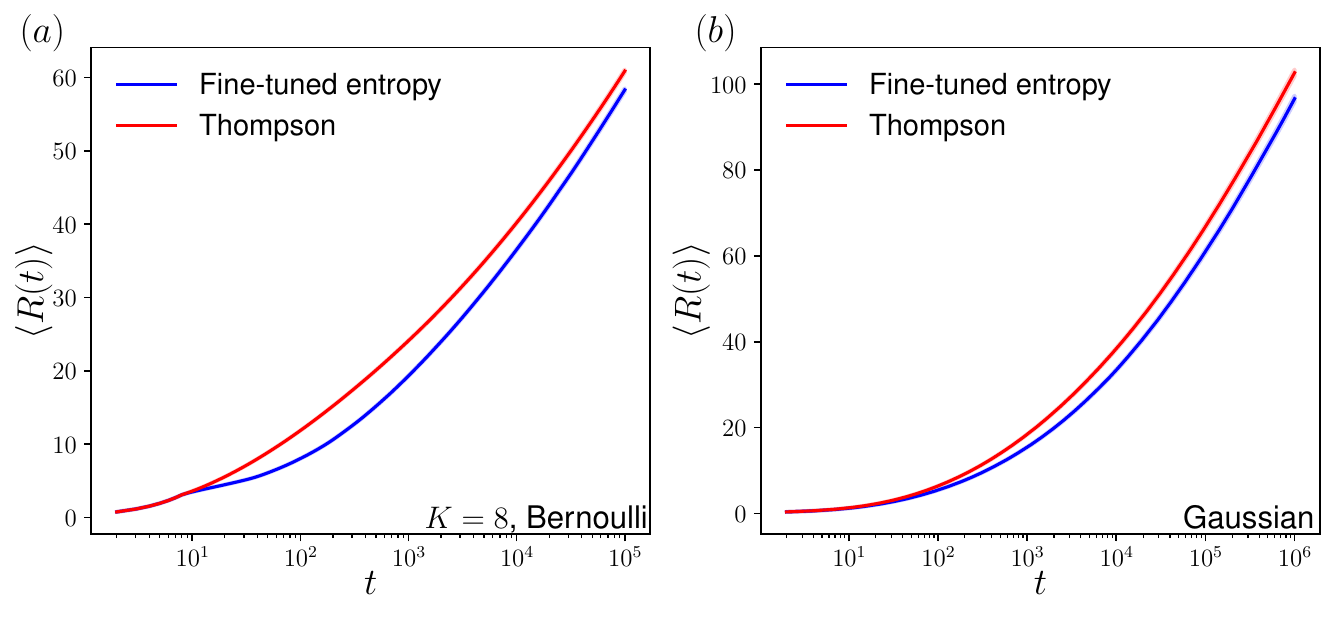}
\caption{Mean regret for 8-armed Bernoulli \textbf{(a)} and Gaussian \textbf{(b)} bandits with parameters drawn uniformly in $]0,1[$. In red Thompson sampling and in blue tuned \algoname (see Section \ref{SItuned}).}
\label{fig:tuned}
\end{figure*}

Here, we detail how the \algoname algorithm can be tuned for specific multi-armed bandit problems, showing its capacity to outperform Thompson sampling (see Fig.~\ref{fig:tuned}).
We propose some empirically optimised variations to the functional form of the approximate entropy $\Sapp$ [Eq.~\eqref{theoryS} and Table~\ref{table:maintable}] derived in the main text. 
These lead to a simplified version of the approximate entropy, which is adjusted to provide a better expected regret in case of expected rewards drawn according to a uniform prior while still respecting the Lai and Robbins bound. 

\subsection{Bernoulli rewards with $K > 2$ }\label{Situnedmulti}

To obtain a tuned version of \algoname for multi-armed bandits, we propose to simplify the main mode term by keeping the dominant term (when $\Delta/\sqrt{2V_t} \to \infty$) which we multiplied by two:

\begin{equation}
\Sc = - \ln \left( 2 \pi  \frac{\meana{\maxa} (1-\meana{\maxa})}{\N{\maxa}}  \right).
\end{equation}

We also simplified the expression by neglecting the contribution from $\ptail$ (i.e., letting $\ptail\to0)$. 
This leads to the tuned expression:
\begin{equation}\label{theoryS_Multiversion}
\begin{split}
\Sapp &\approx  \Sc + \Stail .
\end{split}
\end{equation}

Thus, the tuned version of \algoname for multi-armed Bernoulli bandits consists in replacing $\Sapp$ functional in Eq.~\eqref{Bernoullimulitgrad} and Eq.~\eqref{Bernoullimulitgrad2} by Eq.~\eqref{theoryS_Multiversion}.

\subsection{Gaussian rewards with $K > 2$}
For the multi-armed Gaussian bandits, we propose the following form:
\begin{equation}\label{Situned1}
\begin{split}
&\Sapp (\meana{\maxa},\N{\maxa},\N{\mina},\meana{\mina})  =  \frac{1}{8} \left[1+  \erf \left(  \frac{\teq-\meana{\mina}}{\sqrt{2 \sigma^2 \N{\mina}^{-1}}} \right) \right] \ln \left(\frac{ 2\pi e^{1 -2 \Ac}\sigma^2}{\N{\maxa}} \right)  + 2\ln \left( \frac{ 2\pi e\sigma^2}{\N{\mina}} \right) \erfc \left(\frac{\teq-\meana{\mina}}{\sqrt{2\sigma^2 \N{\mina}^{-1}}} \right). \\
\end{split}
\end{equation}

Since, Eq.~\eqref{Situned1} exhibits a simple closed-form expression, it is possible to derive an exact and explicit expression of its expected gradient for continuous Gaussian reward distributions,
\begin{equation}\label{Situned2}
\begin{split}
\gradsp &=  \int_{-\infty}^{\infty} \frac{e^{-\frac{\mu^2}{2 \sigma^2}}}{\sqrt{2\pi \sigma^2}} \bigg[ | \Sp( \frac{\meana{\maxa} \N{\maxa} + \mu}{\N{\maxa} + 1}, \N{\maxa}+ 1,..) - \Sp(..)| -  | \Sp(..,\frac{ \meana{\mina} \N{\mina}   + \mu}{\N{\mina} + 1}, \N{\mina}+1) - \Sp(..)| \bigg]d\mu\\
&= \int_{-\infty}^{\infty} \frac{e^{-\frac{\mu^2}{2 \sigma^2}}}{\sqrt{2\pi \sigma^2}} \bigg[  |\Delta_{\maxa} \Sp| - |\Delta_{\mina} \Sp| \bigg],
\end{split}
\end{equation}
where the two dots refer to constant variables. Noticing that the first term (variation along $\maxa$) is independent of the integration variable, we obtain:
\begin{equation}\label{Situned2bis}
\begin{split}
\Delta_{\maxa} \Sapp &= -\frac{1}{8} \left[1+  \erf \left( \frac{ \teq-\meana{\mina})}{\sqrt{2\sigma^2 \N{\mina}^{-1}}} \right) \right]  \ln \left(  1 + \frac{1}{\N{\maxa} } \right). \\
\end{split}
\end{equation}
By use of the identity Eq.~\eqref{Sierfint} this leads to:
\begin{equation}\label{Situned4}
\begin{split}
\Delta_{\mina} \Sapp &=   \frac{1}{8}\ln \left( \frac{2\pi e^{1 -2 \Ac}\sigma^2} {\N{\maxa}} \right)   \left[ \erf \left(\frac{ (\teq-\meana{\mina})(\N{\mina} + 1)}{\sqrt{2\sigma^2} \sqrt{\N{\mina} + 2 }} \right) - \erf \left( \frac{\teq - \meana{\mina}}{\sqrt{2 \sigma^2 \N{\mina}^{-1} }} \right) \right] \\
& \hspace{3cm} + 2\ln \left( \frac{2\pi \sigma^2 e^{1}}{ \N{\mina}+1}\right)  \erfc \left( \frac{  (\teq-\meana{\mina})(\N{\mina} + 1)}{\sqrt{2\sigma^2} \sqrt{\N{\mina} + 2 }} \right) - 2\ln \left(\frac{2\pi \sigma^2 e^{1}}{\N{\mina}} \right)  \erfc \left(  \frac{\teq - \meana{\mina}}{\sqrt{2 \sigma^2 \N{\mina}^{-1} }} \right). \\
\end{split}
\end{equation}

Combining \ref{Situned2bis} and \ref{Situned4} leads to the complete expression of the gradient: 
\begin{equation}\label{Situned5}
\begin{split}
&\Delta_{\mathrm{max, min}}\Sapp =  \frac{1}{8} \left[1+  \erf \left( \frac{ \teq-\meana{\mina}}{\sqrt{2\sigma^2 \N{\mina}^{-1}}} \right) \right]  \ln \left(  1 + \frac{1}{\N{\maxa} } \right) \\& \hspace{4cm}- \frac{1}{8}\ln \left( \frac{\N{\maxa}}{2\pi e^{1 -2 \Ac}\sigma^2} \right)   \left[ \erf \left(\frac{ (\teq-\meana{\mina})(\N{\mina} + 1)}{\sqrt{2\sigma^2} \sqrt{\N{\mina} + 2 }} \right) - \erf \left( \frac{\teq - \meana{\mina}}{\sqrt{2 \sigma^2 \N{\mina}^{-1} }} \right) \right] \\
& \hspace{4cm} -2 \ln \left( \frac{ \N{\mina}+1}{2\pi \sigma^2 e^{1}} \right)  \erfc \left( \frac{  (\teq-\meana{\mina})(\N{\mina} + 1)}{\sqrt{2\sigma^2} \sqrt{\N{\mina} + 2 }} \right) + 2\ln \left(\frac{\N{\mina}}{2\pi \sigma^2 e^{1}} \right)  \erfc \left(  \frac{\teq - \meana{\mina}}{\sqrt{2 \sigma^2 \N{\mina}^{-1} }} \right). \\
\end{split}
\end{equation}

Thus, the tuned and continuous version of \algoname for Gaussian rewards consists in replacing gradient evaluation in Eq.~\eqref{gradientgauss} by Eq.~\eqref{Situned5}
\end{widetext}

\end{document}